# GRU-D Characterizes Age-Specific Temporal Missingness in MIMIC-IV

Niklas GIESA[a,1], Mert AKGUEL[a], Sebastian Daniel BOIE[a], and Felix BALZER[a]

[a] *Institute of Medical Informatics, Charité - Universitätsmedizin Berlin, 10117 Berlin*

**Abstract.** Temporal missingness, defined as unobserved patterns in time series, and its predictive potentials represent an emerging area in clinical machine learning. We trained a gated recurrent unit with decay mechanisms, called GRU-D, for a binary classification between elderly - and young patients. We extracted time series for 5 vital signs from MIMIC-IV as model inputs. GRU-D was evaluated with means of 0.780 AUROC and 0.810 AUPRC on bootstrapped data. Interpreting trained model parameters, we found differences in blood pressure missingness and respiratory rate missingness as important predictors learned by parameterized hidden gated units. We successfully showed how GRU-D can be used to reveal patterns in temporal missingness building the basis of novel research directions.

**Keywords.** GRU-D, MIMIC-IV, ICU, Temporal Missingness

## 1. Introduction

The temporal course of longitudinal vital signs, in form of time series, was found to be important for predicting clinical outcomes, like delirium or sepsis [1,2]. Advanced machine learning (ML) models, like gated recurrent units (GRU) [3], process time series directly, often imputing longitudinal data with the last observed value (LOV). The predictive value of temporal missingness, as patterns of unobserved values in time series, for outcome prediction is currently under-explored.

Tan et al. [4] identified patters in unobserved laboratory values for COVID patients expressing higher missingness for less critically ill patients in intensive care units (ICUs). Yuan et al. [5] claim that clinical studies might be biased by temporal missingness towards the outcome of interest. Kaplan et al. [6] pointed out that irregular sampled clinical time series pose challenges due to data loss via applying equidistance time grids for ML. Che et al. [7] found informative temporal missingness in the 3$^{rd}$ version of the Medical Information Mart for Intensive Care (MIMIC-III) related to diagnose codes. Additionally, the authors Che et al. extended the GRU architecture by so-called decays (GRU-D), explicitly learning temporal missingness patterns. Up to now, GRU-D has gained a lot of popularity in the field of clinical ML.

In this study, we investigate age-specific temporal missingness patterns for ICU patients with the newest open data from MIMIC-IV. The elderly oftentimes suffer from more critical situations, like advanced life support (ALS), than the young. Thus, we hypothesize that older patients have more available data due to an intensified clinical monitoring. For specifically analyzing missingness patterns, we interpret a fitted GRU-D that learned to discriminate between two age groups.

*1.1. MIMIC-IV and Temporal Missingness*

We extracted clinical time series from the open clinical database MIMIC-IV [8]. We selected five vital signs; heart rate (hr), blood oxygen saturation (spo2), respiratory rate (rr), systolic blood pressure (bp_sys), and diastolic blood pressure (bp_dia). We integrated variables into a 1h equidistant time grid [9]. Our cohort definition covered patients with a length of ICU stay (lo-icu) between one - and five days. Due to the investigation of temporal missingness patterns, we set outliers to extreme value ranges instead of removing them, e.g., range of [0, 400] for bp_sys. This cleansing affected about 12% of included data records preserving original missingness. To compare patient characteristics, we used Welch's t-test that tests for equal means [10]. The test can ingest unequally sized samples while being robust against Type I errors [11]. In our study, we further defined temporal time series missingness (*TSM*) as **a)** $TSM = N_{miss} / N$, where $N$ is the number of all values, $N_{miss}$ the number of missing ones within one time series.

*1.2. Baseline Models and GRU-D*

We trained baseline models as linear logistic regression (LR) [12] and non-linear gradient boosted trees (BT) [13]. LR was configured with a $L_1$ penalty term (strength C = 0.1), a regularization technique that shrinks non-important coefficients to zero [14]. The ensemble-learning technique BT comprised 3,000 estimators and a maximum branch depth of 1 allowing tree pruning [15]. Baseline models were trained on tabular data instead of time series. Hence, longitudinal data was aggregated by mean, standard deviation (SD), and quartiles (1st, 2nd, 3rd) [16], added to the *TSM* rate per time series.

Che et al. [7] improved standard GRU gates by the hidden decay (D[h]) - and the input decay ([D[x]) mechanism. As an intuition, unobserved values are assumed to converge towards their observed means when missing over time. Both mechanisms (D[h], (D[x]) were realized by trainable decay rates as **b)** $\gamma_t = \exp\{-\max(0, W_\gamma b_t + b_\gamma)\}$, with $W$ and $b$ being learned jointly with all other parameters at the *t*-th timestep. Values of $\gamma_t$ range between 0 and 1 where a deviation from 1 expresses a stronger asymptotic behavior towards the mean than relying on the LOV.

GRU-D's feature space comprises **1.** Input time series values (X), **2.** Binary missing indicators (BMI) as masks (1=missing, 0=present at *t*), **3.** Time intervals as deltas $\Delta_t$ describing how much time passed between the LOV and the present value, **4.** All LOVs within time series. Additionally, the model is fed with means from the training set to model the convergence behavior. All standard GRU gates (update, hidden, reset) were of dimension 5, determined by the number of features (see [7] for details).

*1.3. Training and Validation*

All ML models learned to minimize a binary cross entropy (BCE) [17] loss, eligible for classifications. We split all data according to patient ids into train (70%) and test (30%) sets, avoiding estimation bias. Both sets were z-transformed [18] using means and SDs from the train set only. GRU-D was trained with batches of 64 and a learning rate of 1E-4 for 40 epochs. For validations, the test set was 100x bootstrapped as resampling with replacement [19], enabling the calculation of confidence intervals (CIs).

[1]MAIL: niklas.giesa@charite.de, ORCID: https://orcid.org/0000-0003-0808-3966

## 2. Results

### 2.1. Patient Characteristics

| Variable | All Patients | Patients < 65.0 Years | Patients ≥ 65.0 Years | P-Value |
|---|---|---|---|---|
| **Counts** | | | | |
| #subjects | 24,154 | 11,411 | 12,743 | - |
| #icu-stays | 28,860 | 13,981 | 14,879 | - |
| #records | 1,744,246 | 833,578 | 910,668 | - |
| **General Information** | | | | |
| sex (female/male) [%] | (43/57) | (42/58) | (46/53) | - |
| age [years] | 64.0, [54.0, 65.0, 76.0] | 49.0, [42.0, 52.0, 59.0] | 76.0, [69.0, 75.0, 82.0] | - |
| lo-icu [days] | 2.32, [1.41, 2.01, 2.97] | 2.31, [1.39, 1.97, 2.91] | 2.33, [1.42, 2.0, 3.0] | 4.67E-07 |
| lo-seq [hours] | 60.44, [35.0, 49.0, 74.0] | 59.46, [34.0, 48.0, 73.0] | 61.28, [36, 50, 76] | 1.55E-09 |
| **Temporal Missingness** | | | | |
| hr TSM [%] | 10.25, [2.42, 5.17, 11.68] | 11.75, [2.48, 5.65, 13.47] | 8.94, [2.41, 4.84, 10.13] | 3.60E-02 |
| rr TSM [%] | 83.72, [76.0, 84.48, 92.44] | 83.71, [76.64, 84.31, 92.47] | 83.72, [75.54, 84.61, 92.39] | 1.49E-03 |
| spo2 TSM [%] | 23.48, [7.1, 17.75, 34.96] | 25.73, [8.59, 21.33, 38.84] | 20.84, [5.87, 13.64, 30.35] | 1.32E-17 |
| bp_sys TSM [%] | 31.71, [11.39, 26.29, 46.62] | 32.66, [12.29, 27.61, 47.77] | 30.83, [10.68, 25.41, 45.88] | 8.47E-15 |
| bp_dia TSM [%] | 31.76, [11.36, 26.43, 46.75] | 32.53, [11.98, 27.47, 47.70] | 31.04, [10.82, 25.61, 46.07] | 6.95E-15 |

**Table 1**: Patient characteristics of study cohort for all, younger - (age < 65.0 years), and elderly (age ≥ 65.0 years) patients. Distributions are reported as mean, [1st, 2nd, 3rd quartile]. P-values are derived from Welch's t-test comparing age groups. We report the length of ICU (lo-icu) and – sequences (lo-seq) in addition to time series missingness (TSM) for 5 vital signs.

We report cohort characteristics (see **Table 1**). In the median, patients were 65 years old. Thus, we divided our cohort in younger - (age < 65.0 years, $y$=0) and elderly (age ≥ 65.0 years, $y$=1) patients for the detection of age differences with $y$ as dependent variable. Older patients stayed significantly longer in the ICU (p-value = 5.67E-07) with extended time series (p-value = 1.55E-09). The rr values were in general highly missing, hr records were mostly present. TSM rates were higher for younger - than for elderly patients across all vitals, except for rr. Blood oxygen (spo2) expressed the strongest effect (means = 25.73/20.84, p-value = 1.32E-17).

### 2.2. Model Fits

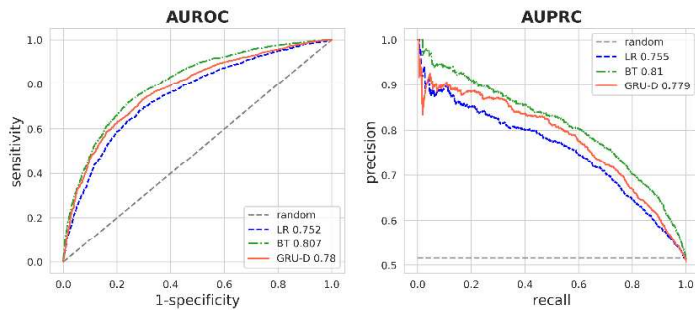

**Figure 1**: Area under (AU-) receiver operating characteristics (-ROC) or precision recall (-PRC) curves for logistic regression (LR), boosted trees (BT), decayed gated recurrent unit (GRU-D) predicting elderly class (≥ 65.0 Years) with first 24h in ICU.

We trained all models on the first 24h of time series in the ICU to classify $y$. We evaluated model fits with AUROC and AUPRC scores [20] (see **Figure 1**). GRU-D achieved an AUROC of 0.780 [0.777, 0.791] (mean, 95%-CI) and an AUPRC of 0.810 [0.796, 0.811]. The model could outperform LR (AUROC of 0.752 [0.743, 0.761], AUPRC of 0.755 [0.744, 0.759]), but performances did not exceed BT scores (AUROC of **0.807** [0.795, 0.805], AUPRC of **0.810** [0.796, 0.811]). While AUROC curves did not overlap across models, LR and GRU-D curves intersected around 1E-2 recall for the AUPRC metric (see **Figure 1**).

[1]MAIL: niklas.giesa@charite.de, ORCID: https://orcid.org/0000-0003-0808-3966

*2.3. Decay Rates*

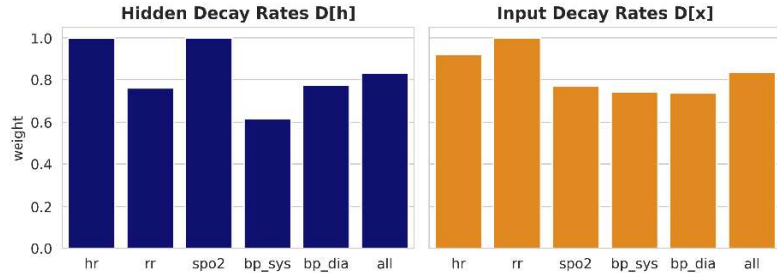

**Figure 2**: Mean weights of hidden- (D[h] left graph) and input (D[x] right graph) decay rates (y-axis) per single feature and all features (x-axis) averaged across all 24 timesteps (=1h of time series) and across all patients.

We read and mean-aggregated decay rates for D[h] and D[x] across time steps per feature. In the long-term, GRU-D learned especially from temporal missingness of bp_sys, bp_dia, and rr recordings (see left graph in **Figure 2**). The input decay received more temporal missing signals from other features than rr (see right graph in **Figure 2**).

When averaging across features per single time step, long-term missingness between 23-24h after admission (D[h] ca. 82.97E-2) was more important for age discrimination than at other time points (D[h] ca. 83.00E-2). In the short-term, GRU-D focused its D[x] rates on 2-4h (D[x] ca. 83.41E-2) than other hours after admission (D[x] ca. 83.42E-2).

Overall, we detected slightly more signal in D[h] decays (mean of 82.99E-2) than in D[x] (mean of 83.41E-2) across all features, patients, and time steps.

## 3. Discussion

This study successfully detected age-specific differences in temporal missingness for a large openly available patient cohort. GRU-D yielded good validation performances with a mean AUROC of 0.78. Complex model parameters must be trained that might have led to the superior performance (mean AUROC of 0.807) of ensemble models (BT) over GRU-D. Vital signs were missing more often within sequences (TSM) for patients younger than 65 years. Elderly could have required a narrower bedside monitoring due to more severe conditions. Results align with findings from Yuan et al. [4] indicating an increase in data availability with an increase of illness severity.

Specific high TSM of rr recordings (>80%) could be caused by invasive ventilation procedures, also leading to many empty time series and low input D[x] decay rates. However, GRU-D efficiently learned with these small present signals in the long-term gates (D[h]). Both blood pressure types (bp_sys, bp_dia) contributed equally to learnable D[x] decays, while GRU-D learned different long-term temporal missingness. Here, the absence of one or another type could have discriminative value for learned hidden decays D[h]. Differences might be caused by erroneous bp monitoring for critical situations like ALS. Coefficient strengths of baseline LR were elevated for spo2 TSM with 0.115 alongside blood pressure summary statistics ranging between 0.125 and 0.411 highlighting the importance of temporal missingness for age discrimination.

[1]MAIL: niklas.giesa@charite.de, ORCID: https://orcid.org/0000-0003-0808-3966

Similar to Che et al. [7] that found informative missingness in MIMIC-III, we could also observe these patterns in the newest version of the open database aka MIMIC-IV. As expected, the GRU-D model stored the most long-term signals toward the end of the first 24h in the ICU. On the contrary, short-term input signals were especially learned in the beginning of the ICU stay, e.g., due to attaching a patient initially to clinical monitors.

Our work is limited by selecting time series data from one single-center database. Thus, results must be validated further to claim generalizability. We trained models with randomly initialized parameters incrementally changing them to improve performances. Sophisticated search methods, like Hyperband, could have improved our models.

We see GRU-D as a sophisticated method for jointly learning clinical outcomes and missingness. Further extensions, like ordinary differential equations for uncertainty estimations suggested by Habiba et al. [21], seem to be a promising future research field.

## 4. Conclusion

This work showed age-related differences in temporal missingness patterns described by an advanced ML model named GRU-D on the large-scale open dataset MIMIC-IV.

[1]MAIL: niklas.giesa@charite.de, ORCID: https://orcid.org/0000-0003-0808-3966